\documentclass[runningheads]{llncs}

\usepackage{eccv}

\usepackage{eccvabbrv}

\usepackage{graphicx}
\usepackage{booktabs}
\usepackage{caption}

\usepackage[accsupp]{axessibility}  % Improves PDF readability for those with disabilities.

\usepackage{hyperref}

\usepackage{orcidlink}

\begin{document}

\title{SC4D: Sparse-Controlled Video-to-4D Generation and Motion Transfer} 

\titlerunning{SC4D: Sparse-Controlled Video-to-4D Generation and Motion Transfer}

\renewcommand{\thefootnote}{}
\footnotetext[2]{
* Work done during Zijie's internship at DAMO Academy, Alibaba Group \\
$^\dag$ Corresponding author}

\author{Zijie Wu\inst{1,2} \and
Chaohui Yu\inst{2,3} \and
Yanqin Jiang\inst{2} \and
Chenjie Cao\inst{2,3} \and
Fan Wang\inst{2,3} \and
Xiang Bai\inst{1}$^\dag$
}

\authorrunning{Wu et al.}

\institute{Huazhong University of Science and Technology \and
DAMO Academy, Alibaba Group
\and
Hupan Lab \\
\email{\{zjw1031,xbai\}@hust.edu.cn, \{huakun.ych,jiangyanqin.jyq,caochenjie.ccj,fan.w\}@alibaba-inc.com}\\
\url{https://sc4d.github.io/}
}

% \maketitle

%Add a sexy teaser
\makeatletter
\let\@oldmaketitle\@maketitle
\renewcommand{\@maketitle}{\@oldmaketitle
 \centering
    \includegraphics[width=1.0\linewidth]{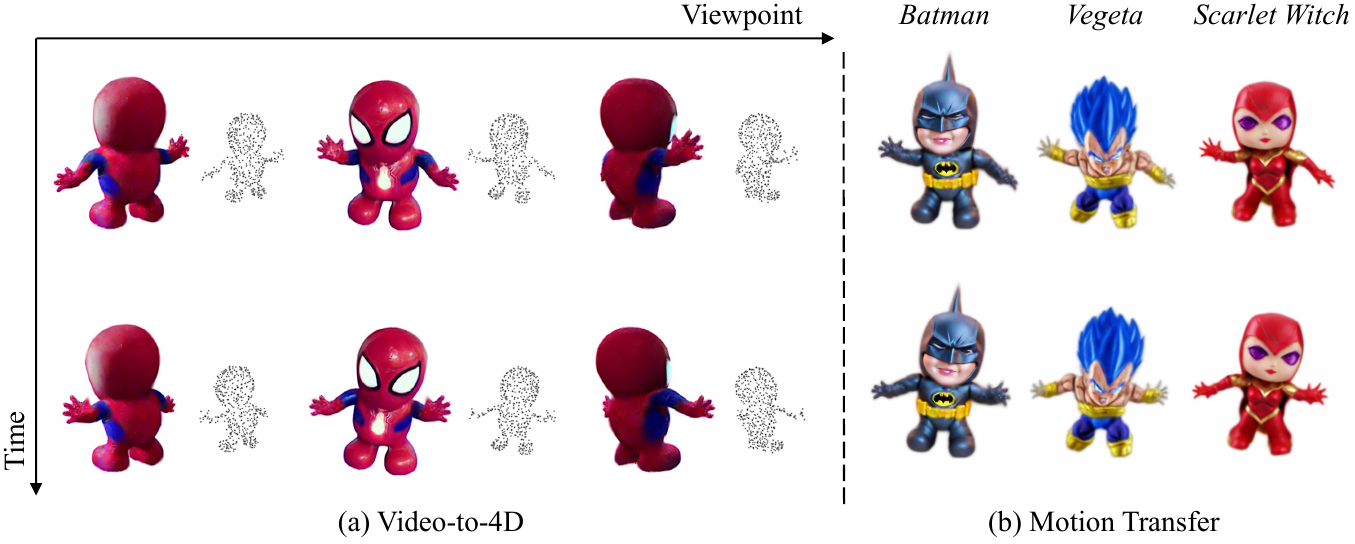}
    \vspace{-.05in}
    \captionof{figure}{We illustrate: (a) video-to-4D results of SC4D and corresponding control points visualizations, and (b) examples of our motion transfer applications in the figure.}
    \label{fig:teaser}
  \bigskip}
\makeatother

\maketitle

% \newcommand\blfootnote[1]{%
% \begingroup
% \renewcommand\thefootnote{}\footnote{#1}%
% \addtocounter{footnote}{-1}%
% \endgroup
% }

% \blfootnote{$\dagger$ Corresponding author}

% \vspace{-.2in}

\begin{abstract}
  Recent advances in 2D/3D generative models enable the generation of dynamic 3D objects from a single-view video. Existing approaches utilize score distillation sampling to form the dynamic scene as dynamic NeRF or dense 3D Gaussians. However, these methods struggle to strike a balance among reference view alignment, spatio-temporal consistency, and motion fidelity under single-view conditions due to the implicit nature of NeRF or the intricate dense Gaussian motion prediction. To address these issues, this paper proposes an efficient, sparse-controlled video-to-4D framework named SC4D, that decouples motion and appearance to achieve superior video-to-4D generation. Moreover, we introduce Adaptive Gaussian (AG) initialization and Gaussian Alignment (GA) loss to mitigate shape degeneration issue, ensuring the fidelity of the learned motion and shape. Comprehensive experimental results demonstrate that our method surpasses existing methods in both quality and efficiency. In addition, facilitated by the disentangled modeling of motion and appearance of SC4D, we devise a novel application that seamlessly transfers the learned motion onto a diverse array of 4D entities according to textual descriptions.
  %Moreover, leveraging the control points gained by the SC4D, we propose a novel application that seamlessly transfers the learned motion onto 4D objects of different identities according to textual descriptions.
  %Project page~\footnote{SC4D project page:~\href{https://sc4d.github.io/}{\textcolor{red}{https://sc4d.github.io/}}}. 

\end{abstract}

\section{Introduction}
\label{intro}

% Inferring geometry and appearance of 3D objects from a single image is one of the most fundamental and challenging task in computer vision, which is critical for many downstream applications, such as AR/VR, 3D printing, 3D simulations. Due to the difficulty of maintaining temporal consistency and restoring accurate motion, recovering dynamic objects from a single   video is even more formidable. However, as humans, we are adept at resolving the above ill-posed tasks, a capability attributable to our possession of extensive prior knowledge of the real world.

In recent years, with the advancement of generative AI, we have witnessed significant progress in 3D generation techniques, which include the generation of static objects' shape, texture, and even an entire scene from a text prompt or a single image. Compared to static 3D assets, dynamic 3D (4D) content offers greater spatio-temporal flexibility and thus harbors more substantial potential for applications in AR/VR, filming, animation, simulation, and other domains. However, generating 4D objects from text descriptions or video references is rather formidable due to the difficulties of maintaining spatio-temporal consistency and ensuring motion fidelity. Nevertheless, as humans, we are adept at resolving the above challenging tasks, a capability attributable to our possession of extensive prior knowledge of the real world.

Very recently, building upon the foundations laid by extant text-to-3D~\cite{dreamfusion, sjc, magic3d, points-to-3d, prolificdreamer} and image-to-3D~\cite{make-it-3d, magic123, one2345, consistent123, dreamcraft3d} pipelines, several methods~\cite{con4d, 4dgen, efficient4d} distill prior knowledge from novel view synthesis models~\cite{zero123, syncdreamer} to generate the target 4D object as dynamic NeRF~\cite{con4d} or dynamic 3D Gaussians~\cite{4dgen, efficient4d} and impose constraints to ensure the temporal consistency among frames. Despite the commendable progress achieved by these methods, they still struggle to strike a balance among reference view alignment, spatio-temporal consistency, and motion fidelity. We argue that the representation is critical for video-to-4D generation. On the one hand, dynamic NeRF~\cite{nerf, dnerf, hexplane, kplanes} cannot sustain temporal coherence under novel views without additional restrictions due to its implicit nature and stochasticity inherent in Score Distillation Sampling (SDS)~\cite{dreamfusion}. On the other hand, it is intricate to predict accurate trajectories and rotations for dense Gaussians~\cite{3dgs, dynamic3dgs, 4dgs, deformable3dgs} with only single-view conditions.

To tackle the aforementioned problems, inspired by a recent dynamic scene reconstruction method~\cite{scgs}, we decouple the appearance and motion of the dynamic 3D object into dense Gaussians ($\sim$50k) and sparse control points ($\sim$512) and design an efficient two-stage video-to-4D generation framework, named~\textbf{SC4D}. Specifically, in the coarse stage, we initialize a set of sparse control points as spherical Gaussians and a Multilayer Perceptron (MLP) conditioned on time and location to predict the motion of these sparse Gaussians. Then, we optimize the parameters of these Gaussians and the MLP under the guidance of reference view reconstruction and novel view score distillation. In the fine stage, we utilize the sparse control Gaussians as implicit control points and perform Linear Binding Skinning (LBS)~\cite{lbs} to drive the dense Gaussians. In this stage, we jointly optimize the parameters of control points, dense Gaussians, and deformation MLP to obtain the final results. Notably, since only a single-view ground truth video is provided, we empirically find that there is a proclivity for shape degeneration issues in the fine stage, which results in thickening, position displacements, and texture blur of the dynamic 3D object. To address these challenges, we devise Adaptive Gaussian (AG) initialization that inherits the shape and motion of the control Gaussians in the coarse stage with a random amount of Gaussians. Moreover, we present Gaussian Alignment (GA) loss to ensure shape and motion fidelity in the fine stage.   

Benefiting from the disentangled modeling of appearance and motion within our method, SC4D effectively mitigates the ambiguity of these two attributes during optimization. Additionally, since the motion of the dynamic object is governed by a set of sparse control points rather than dense Gaussians, our approach simplifies the learning of motion and naturally exhibits enhanced local rigidity. Comprehensive evaluations reveal that our method surpasses existing video-to-4D generation methods~\cite{con4d, 4dgen} in both quality and efficiency. Moreover, after obtaining the motions of implicit control points, we introduce a novel application that transfers the learned motion onto other entities under the guidance of text-to-image models~\cite{sd, controlnet} and dense Gaussians' depth. In conclusion, our contributions can be summarized as follows:

\begin{enumerate}
    \item We propose SC4D, a sophisticated video-to-4D generation framework based on sparse points control, which generates dynamic 3D objects with superior quality and efficiency compared to existing methods.
    \item We devise an Adaptive Gaussian (AG) initialization approach and Gaussian Alignment (GA) loss that effectively mitigate the shape degeneration issue, ensuring accurate motion and shape learning.
    \item We propose a novel application based on the control points' motion and design a motion transfer pipeline that 
    maps the learned motion onto distinct entities, as directed by textual descriptions.
\end{enumerate}

\section{Related Works}
\label{related_works}

\textbf{Optimization-based 3D Generation.}
Optimization-based 3D generation methods typically optimize a NeRF~\cite{nerf} or 3D Gaussians~\cite{3dgs} utilizing prior knowledge from image-text matching model~\cite{clip} or diffusion-based generative models~\cite{ddpm, sd, imagen, zero123}. 
%According to the type of input sources, these methods can be roughly divided into two categories: text-to-3D and image-to-3D.
CLIP-based text-to-3D methods~\cite{dreamfield, clip-mesh, pureclipnerf, dream3d} generally employ CLIP~\cite{clip} to align each viewpoint of the target 3D scene with the given text description for optimization. 
%Since diffusion-based text-to-image models~\cite{sd, imagen} have shown their sophistication in generating exquisite images, 
DreamFusion~\cite{dreamfusion} substitutes the guidance model from CLIP to a 2D diffusion model~\cite{imagen}, and introduces Score Distillation Sampling (SDS) to distill prior knowledge from text-to-image models. 
%Owing to its elevated generation quality and versatility, the paradigm predicated on SDS has emerged as the mainstream in subsequent works. 
As a concurrent work, SJC~\cite{sjc} shares a similar idea, which distills scores in a Perturb-and-Average manner. 
%Magic3D~\cite{magic3d} incorporates InstanceNGP~\cite{instantngp} and DMTet~\cite{dmtet} in a coarse-to-fine scheme, allowing high-resolution 3D generation without sacrificing efficiency. Fantasia3D~\cite{fantasia3d} introduces the PBR material model~\cite{pbr} and disentangles geometry and appearance for vivid results. Points-to-3D~\cite{points-to-3d} and 3DFuse~\cite{3dfuse} alleviate the Janus problem of 3D generation with Point-E~\cite{pointe} condition. Inspired from VAE~\cite{vae}, ProlificDreamer~\cite{prolificdreamer} presents Variational Score Distillation (VSD), which brings the texture fidelity to a new level. 
Following the paradigm of SDS, a series of methods aim at bringing finer texture details~\cite{magic3d, fantasia3d, prolificdreamer} or alleviating the Janus problem~\cite{points-to-3d, 3dfuse} utilizing DMTet~\cite{dmtet} representation, Variational Score Distillation~\cite{prolificdreamer}, Point-E~\cite{pointe} condition, etc. 
Recently, a succession of methods further enhanced the view quality~\cite{hifa, sweetdreamer} and multi-view consistency~\cite{mvdream}. There are also several 3D Gaussian-based methods~\cite{dreamgaussian, gsgen, gaussiandreamer} that achieve comparable results.
As for image-to-3D, RealFusion~\cite{realfusion} performs textual inversion~\cite{ti} before 3D generation to match the intended concept. Make-it-3D~\cite{make-it-3d} improves the view quality and consistency in an inpainting manner. Zero123~\cite{zero123} utilizes large-scale multi-view data from Objaverse dataset~\cite{objaverse, objaverse-xl} to turn Stable Diffusion (SD)~\cite{sd} into a novel-view generator. Based on Zero123, a bunch of methods~\cite{magic123, one2345, dreamcraft3d} achieves high-fidelity image-to-3D generation. There are also several approaches~\cite{consistent123, instant3d, syncdreamer, imagedream} acquire notable progress in enhancing multi-view consistency.  

\noindent\textbf{4D Representation.}
Current 4D representations predominantly bifurcate into two principal categories: dynamic NeRF~\cite{nerf} and dynamic 3D Gaussians~\cite{3dgs}. Dynamic NeRF-based methods can be further divided into deformable NeRF~\cite{dnerf, nerfies, d2nerf, nonrigidnerf} and time-varying NeRF~\cite{dynamicnerf, sceneflow, tineuvox, hexplane, kplanes}. Recently, a variety of methods predicated on dynamic 3D Gaussians~\cite{dynamic3dgs, 4dgs, deformable3dgs, 4dgaussian, gaufre, spacetime3dgs, gssurvey} have emerged, leveraging Gaussians' explicit nature and real-time rendering capabilities. There are also some methods that ameliorate the dense motion prediction by learning a set of sparse trajectories~\cite{dynmf}, control points~\cite{scgs}, or basis vectors~\cite{npgs}. 

\noindent\textbf{4D Generation.}
Compared to 3D generation, high-quality 4D generation is even more challenging since the temporal dimension is involved. Existing text-to-4D methods~\cite{mav3d, 4dfy, animate124, dreamgaussian4d, alignyg, animatabledreamer, 4dscene} distill geometry and temporal information from diffusion-based text-to-image models~\cite{sd} and text-to-video models~\cite{mav} with SDS~\cite{dreamfusion}. In recent developments, several video-to-4D frameworks~\cite{con4d, 4dgen, efficient4d} have been introduced. These pipelines endeavor to recover the dynamic 3D objects from single-view video inputs, facilitated by Zero123~\cite{zero123}. However, these video-to-4D methods struggle to strike a balance among reference view alignment, spatio-temporal consistency, and motion fidelity.

% A recent work pertinent to our application is AnimatableDreamer~\cite{animatabledreamer}, which relies on BANMo~\cite{banmo} to extract bones and time-varying articulations from a single video. However, BANMo cannot handle dynamic objects that lack multi-view information, and it also necessitates a higher number of input frames. In contrast, our method is more general; it is capable of learning the motion represented by control points even when dealing with dynamic objects from a single viewpoint.

\section{Method}
\label{method}

\begin{figure}[t]
\begin{center}
\includegraphics[width=0.95\textwidth]{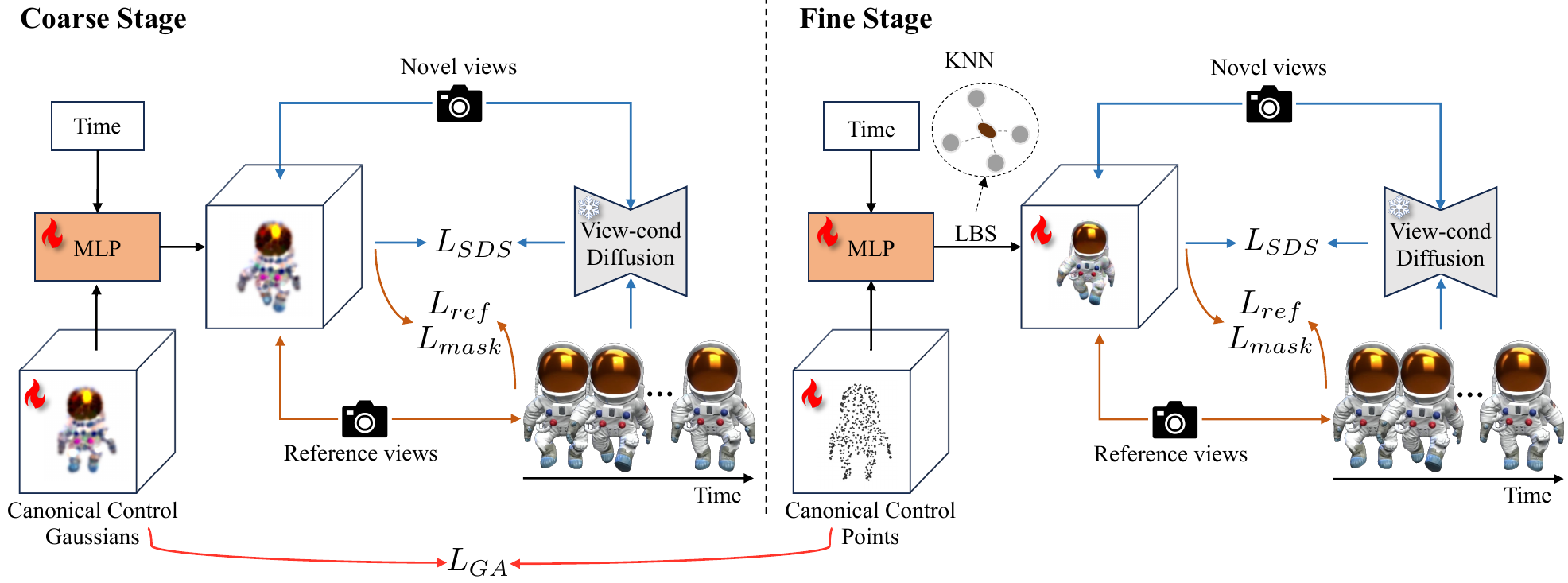}
\end{center}
% \vspace{-.15in}
\caption{Overall pipeline of the proposed SC4D. In the coarse stage, SC4D learns a proper shape and motion initialization with a set of sparse control Gaussians. Then, in the fine stage, we propose Adaptive Gaussian (AG) initialization, and Gaussian Alignment (GA) loss to prevent shape and motion degeneration, and jointly optimize control points, dense Gaussians, and deformation MLP for the final results.}
% \vspace{-.2in}
\label{fig:pipeline}
\end{figure}

%The scarcity of authentic 4D data renders the task of generating a 4D object from a single-view video exceedingly challenging. Current methods~\cite{con4d, 4dgen, efficient4d} adopts dynamic NeRF~\cite{kplanes} or dense dynamic 3D Gaussians~\cite{4dgs} to represent the learned objects in the 4D domain. Nonetheless, it has been observed that these 4D representations struggle to achieve an equilibrium among reference view alignment, spatio-temporal consistency, and motion fidelity under single-view condition. 
Given a single-view reference video, video-to-4D methods~\cite{con4d, 4dgen, efficient4d} aim to recover a plausible dynamic 3D object that aligns with the video source. Inspired by~\cite{scgs}, we propose an efficient video-to-4D framework based on sparse control points (shown in Fig.~\ref{fig:pipeline}), named \textbf{SC4D}, which utilizes separated modeling of appearance and motion to yield superior outcomes. To ensure the fidelity of learned shape and motion, we introduce Adaptive Gaussian (AG) initialization based on control points, and Gaussian Alignment (GA) loss as an additional constraint. Furthermore, we devise a novel application that enables motion transfer with text descriptions after acquiring the control point motions.

% In the following sections, we first introduce the relevant preliminaries in Sec.~\ref{preliminaries}. Then, we illustrate the details of SC4D in Sec.~\ref{coarse} and Sec.~\ref{fine} for the coarse stage and the fine stage, respectively. At last, we describe our motion transfer pipeline in Sec.~\ref{motion_transfer}.

\subsection{Preliminaries}
\label{preliminaries}

\noindent\textbf{Score Distilltion Sampling.}
% Score Distillation Sampling (SDS)~\cite{dreamfusion} is widely adopted to distill prior knowledge from text-to-image models~\cite{sd} or novel view synthesis models~\cite{zero123} for text-to-3D or image-to-3D generation. In this work, we mainly focus on video-to-4D generation from a single-view reference video, which utilizes Zero123~\cite{zero123} as the prior knowledge source of novel view information. Given a reference image, a relative camera extrinsic $\mathrm{(R,T)}$ between the queried view and the input view, Zero123 generates the plausible novel view rendering. We denote the 3D representation model as $\theta$, Zero123 model as $\phi$, reference image as $\mathrm{I_{r}}$, then the SDS loss can be formulated as:
Score Distillation Sampling (SDS)~\cite{dreamfusion} is widely adopted to distill prior knowledge from image generation models~\cite{zero123, sd}. In this work, we utilize Zero123~\cite{zero123} as the source of novel view information. Given a reference image $\mathrm{I_{r}}$, a relative camera extrinsic $\mathrm{(R,T)}$ between the queried and input views, the 3D model as $\theta$, Zero123 as $\phi$, then the SDS loss is as follows:

\begin{equation}
\label{equ1}
\bigtriangledown _{\theta}L_{SDS}(\phi, x) = \mathbb{E}_{t,\epsilon }[\omega (t)(\hat{\epsilon }_{\phi}(z_t; \mathrm{I_r,R,T},t) -\epsilon)  \frac{\partial x}{\partial \theta }],
\end{equation}
where $t$ is the randomly sampled timestep in the diffusion process, $x$ is the rendered image, and $\omega(t)$ is a weighting function depending on the timestep $t$. 
%Intuitively, SDS loss computes the score function of the diffusion model through noisy renderings from random viewpoints, which is utilized as the update direction for the target 3D representation model.

\noindent\textbf{3D Gaussian Splatting.} 
3D Gaussian Splatting (3DGS)~\cite{3dgs} represents a scene as a set of explicit 3D Gaussians. Each Gaussian $G$ has a center position $\mu$, a rotation quaternion $q$, a scaling parameter $s$, opacity $\sigma$ and sphere harmonic (SH) coefficients $sh$. It can be defined as $G(x)=e^{-\frac{1}{2}(x-\mu)^T\sum^{-1}(x-\mu)}$,
%
% \begin{equation}
% \label{equ2}
% G(x)=e^{-\frac{1}{2}(x-\mu)^T\sum^{-1}(x-\mu)},
% \end{equation}
where $\sum$ is the 3D covariance matrix, calculated by $\sum=RSS^TR^T$ ($R,S$ is equivalent to $q,s$). The color of a pixel $u$ is rendered using $\alpha$-blending:
\begin{equation}
\label{equ3}
Color(u)=\sum_{i}SH(sh_i,v_i)\alpha_i\prod_{j=1}^{i-1}(1-\alpha_j),
\end{equation}
where $\alpha_i=\sigma_iG(u)$, $v_i$ is the view direction, and $SH$ is the spherical harmonic function. To enhance the accuracy of fitting across diverse scenes, densification and pruning are adopted based on gradient accumulation during optimization.

\noindent\textbf{Sparse-Controlled Gaussian Splatting (SC-GS).}
% Sparse-Controlled Gaussian Splatting (SC-GS)~\cite{scgs} is an effective dynamic scene reconstruction method, which decouples the appearance and motion of a 4D scene as dense 3D Gaussians and moving sparse control points. For each control point, SC-GS learns a translation vector $T_{i}^{t}\in \mathbf{SE}(3)$, and a rotation matrix $R_{i}^{t}\in \mathbf{SE}(3)$. SC-GS utilizes a time-condition MLP $\mathbf{\Psi}$ to predict $T_{i}^{t}$ and $R_{i}^{t}$. 
SC-GS~\cite{scgs} is an effective dynamic scene reconstruction method, which decouples the appearance and motion of a 4D scene as 3D Gaussians and control points. SC-GS utilizes a time-condition MLP $\mathbf{\Psi}$ to predict the translation $T_{i}^{t}$ and rotation $R_{i}^{t}$ for each control point. 
%given the control point location $p_{i}$:
%
%\begin{equation}
%\label{equ4}
%\mathbf{\Psi}: (p_i, t) \to (T_{i}^{t}, R_{i}^{t}).
%\end{equation}
%For practical implementation, the rotation matrix $R_{i}^{t}$ is calculated by a quaternion $r_{i}^{t}$, which is directly predicted by $\mathbf{\Psi}$. 
% Then, 3D Gaussians are driven by these control points following LBS~\cite{lbs}. For each Gaussian $G_j: (\mu_j,q_j,s_j,\sigma_j,sh_j)$, the warped location $\mu_j^t$ and rotation $q_j^t$ can be computed as a weighted sum of its KNN control points $M_j$:
Then, 3D Gaussians are driven by control points following LBS~\cite{lbs}. For each Gaussian $G_j$, the warped location $\mu_j^t$ and rotation $q_j^t$ can be computed as a weighted sum of its KNN control points $M_j$:
\begin{equation}
\label{equ5}
\mu_j^t = \sum_{k\in M_j}\text{w}_{jk}(R_k^t(\mu_j-p_k)+p_k+T_k^t),
\end{equation}
\begin{equation}
\label{equ6}
q_j^t = (\sum_{k\in M_j}\text{w}_{jk}r_k^t)\otimes q_j,
\end{equation}
where $\text{w}_{jk}$ is a weighting ratio depending on the distance $d_{jk}$ between Gaussian $G_j$ center and its neighboring control point $p_k$, and a learned control radius $o_k$. $p_k$, $r_k^t$ denote the position and rotation quaternion for the control point.
%
% \begin{equation}
% \label{equ7}
% \text{w}_{jk}=\frac{\hat{\text{w}}_{jk}}{\sum_{k\in N_j}\hat{\text{w}}_{jk}},~\mathrm{where}~ \hat{\text{w}}_{jk} = exp(-\frac{d^2_{jk}}{2o^2_{k}}).
% \end{equation}

\subsection{Coarse Stage: Sparse Control Points Initialization}
\label{coarse}
As illustrated in SC-GS~\cite{scgs}, sparse control points initialization is critical for decoupling motion and appearance of the dynamic object/scene. Joint optimization of sparse control points and dense Gaussians directly can result in uneven distribution of control points and may even lead to training collapse. In this stage, we aim to obtain a good initialization for sparse control points' locations and motion that align with the reference video.

As shown in Fig.~\ref{fig:pipeline}, since no ground truth 3D data is available, we initialize $M$ control points (denoted as control Gaussians in this stage) as 3D Gaussians with randomly sampled positions inside a sphere. In order to obtain control Gaussians that are more evenly distributed, we constrain them as spheres with the same scaling parameter ($s$). We denote the control Gaussians as $C_i: (p_i,r_i,s,\sigma_i,sh_i)$, and the reference image sequence as $\{I_r^f,f=1,2,\cdots,F\}$, where $F$ stands for the number of frames of the reference video. Then, for a randomly sampled timestep $t=\frac{f-1}{F}$, we predict the control Gaussians' movement using MLP $\mathbf{\Psi}$. To be noted, since the control Gaussians are spherical, we only need to compute their new position as follows:
\begin{equation}
\label{equ8}
p_i^t=p_i+T_i^t,
\end{equation}
where $p_i$ is the position of $C_i$ in the canonical space. After obtaining the deformed object at timestep $t$, we project it from the reference view to get $\hat{I}^t$ following Equ.~(\ref{equ3}), and compute the reconstruction loss as:
\begin{equation}
\label{equ9}
L_{ref}=\left \| \hat{I}^t - I_r^f \right \| _2^2.
\end{equation}

To better leverage the information from the reference images, we additionally introduce a mask loss term:
\begin{equation}
\label{equ10}
L_{mask}=\left \| \alpha^t - \mathbf{M}_r^f \right \| _2^2,
\end{equation}
where $\mathbf{M}_r$ is the foreground mask of the reference image, and $\alpha^t$ is the accumulated opacity obtained during rendering. As for novel view optimization, we sample $B$ viewpoints randomly within the pitch angle range of $-ver$ to $ver$ degrees and the yaw angle range of -180 to 180 degrees, and compute the average SDS loss following Equ.~(\ref{equ1}). The overall objective in this stage is a weighted combination of the above three losses:
\begin{equation}
\label{equ11}
L_{total}=\lambda_{ref}L_{ref}+\lambda_{mask}L_{mask}+\lambda_{SDS} L_{SDS}.
\end{equation}
% where we empirically set $\lambda_{ref},\lambda_{mask},\lambda_{SDS}$ to $5000.0,500.0,1.0$ as default, respectively.

In this stage, we perform densification and pruning following 3DGS~\cite{3dgs} in the first $n$ iterations. Then we sample $M$ (same amount as initialization) control Gaussians utilizing Farthest Point Sampling (FPS)~\cite{pointnet++} and continue training without densification in the remaining procedure.
%Please refer to Supp. for more details. 

\subsection{Fine Stage: Dense Gaussians Optimization}
\label{fine}
After the coarse stage, we get a reasonable estimation for the dynamic 3D object's motion and shape. Then in the fine stage, we aim to optimize the texture details to match the source video and to further refine the motion and shape for better fidelity. To be noted, the explicit control Gaussians in the coarse stage have transitioned to implicit control points in this stage, denoted as $C_i: (p_i,o_i)$, where $o_i$ is the control radius of each control point, initialized with the scaling parameter $s$ of control Gaussians in the coarse stage.

\begin{figure}[t]
\begin{center}
\includegraphics[width=0.95\textwidth]{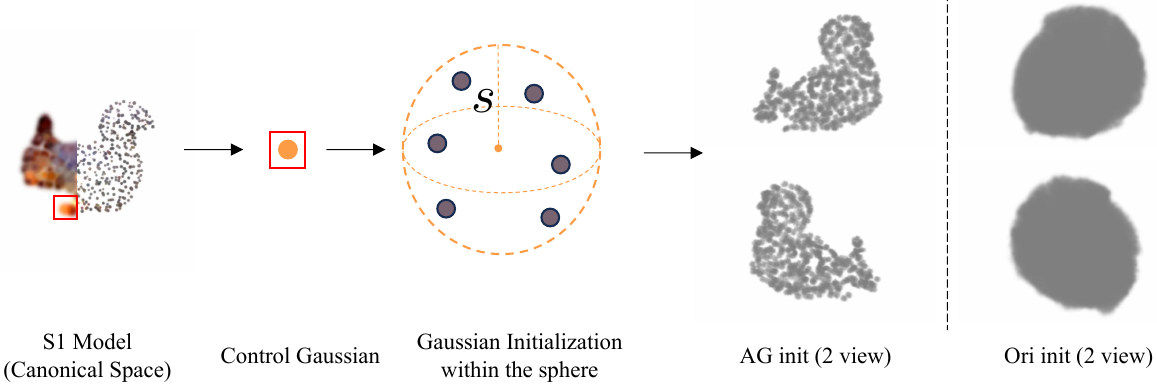}
\end{center}
% \vspace{-.1in}
\caption{Illustration of Adaptive Gaussian (AG) initialization. $s$ is the scaling parameter of control Gaussians in the coarse stage. $Ori~init$ represents randomly initializing all the dense Gaussians within a sphere in the canonical space.}
\label{fig:init}
\end{figure}

\noindent\textbf{Adaptive Gaussian Initialization.} In this stage, dense Gaussians are driven by neighboring sparse control points, as illustrated in Equ.~(\ref{equ5}), (\ref{equ6}). We empirically find that the dense Gaussian initialization significantly impacts the quality of the final result. One straightforward initialization approach is to initialize the dense Gaussians uniformly within a sphere in the canonical space as in the coarse stage. However, we observe that this initialization fails to generate promising results. The target object is prone to increased thickness, and issues such as diminished texture and positional drift may arise (as shown in Fig.~\ref{fig:abl}).

Instead, we propose Adaptive Gaussian (AG) initialization based on the learned control Gaussians. As shown in Fig.~\ref{fig:init}, we have learned $M$ control Gaussians in the coarse stage with the same scaling parameter $s$. Then, for each control Gaussian, we consider it as a sphere with a radius of $s$, and randomly initialize $K$ Gaussians within it following DreamGaussian~\cite{dreamgaussian}. In total, we get $N=M\times K$ Gaussians as initialization. As shown in Fig.~\ref{fig:init}, our designed initialization approach successfully inherits the shape and motion modeled in the coarse stage. The dense Gaussians are distributed near the surface of the object, which facilitates the subsequent optimization. In comparison, if directly optimizing all the dense Gaussians uniformly within a sphere in the canonical space, the deformed shape misaligns with that in the coarse stage.

\noindent\textbf{Gaussian Alignment Loss.}
Even with a good dense Gaussian initialization, the shape and motion of the dynamic 3D object are still prone to degeneration in the later phases of training. The main reason is that: when employing Score Distillation Sampling (SDS)~\cite{dreamfusion} to distill prior knowledge of novel views from Zero123~\cite{zero123}, a larger noise timestep biases SDS towards ensuring the plausibility of the overall shape. Conversely, with a smaller timestep, SDS tends to focus more on optimizing textures, while its capability to preserve shape diminishes.

To allow refining texture without degrading motion and shape, we propose the Gaussian Alignment (GA) loss as an additional constraint. At the beginning of this stage, we preserve the control Gaussians' parameters and the deformation MLP to query the initial positions (denoted as $\overline{p}^t$) of those control points at random timestep $t$. Then, we compute the Gaussian Alignment loss as:
\begin{equation}
\label{equ12}
L_{GA} = \left \| p^t - \overline{p}^t \right \| _2^2,
\end{equation}
where $p^t$ denotes the position of the current control point at timestep $t$. Although the proposed GA loss is quite simple, it can effectively mitigate the shape degeneration issue encountered during the latter training procedure. The Chamfer loss is another commonly used metric for constraining the distances between point clouds. However, compared to GA loss, we observe that the Chamfer loss can sometimes result in the current control points aggregating towards certain target points, thereby compromising the uniform distribution of the control points 
(See Sec.~4.4 of \textit{Supp.}\footnote{\textit{Supp.}: supplementary file.} for the comparison of GA loss and the Chamfer loss). 
% (See Sec.~4.4 of \textit{Supp.} for the comparison of GA loss and the Chamfer loss). 
% (See Sec.~\ref{supp_abl} for the comparison of GA loss and the Chamfer loss).

In this stage, the overall training objective is formulated as follows:
\begin{equation}
\label{equ13}
L_{total}=\lambda_{ref}L_{ref}+\lambda_{mask}L_{mask}+\lambda_{SDS} L_{SDS}+\lambda_{GA} L_{GA}.
\end{equation}
%where $\lambda_{GA}$ is set to $10.0$. Other parameters remain the same as the coarse stage.

\begin{figure}[t]
\begin{center}
\includegraphics[width=0.95\textwidth]{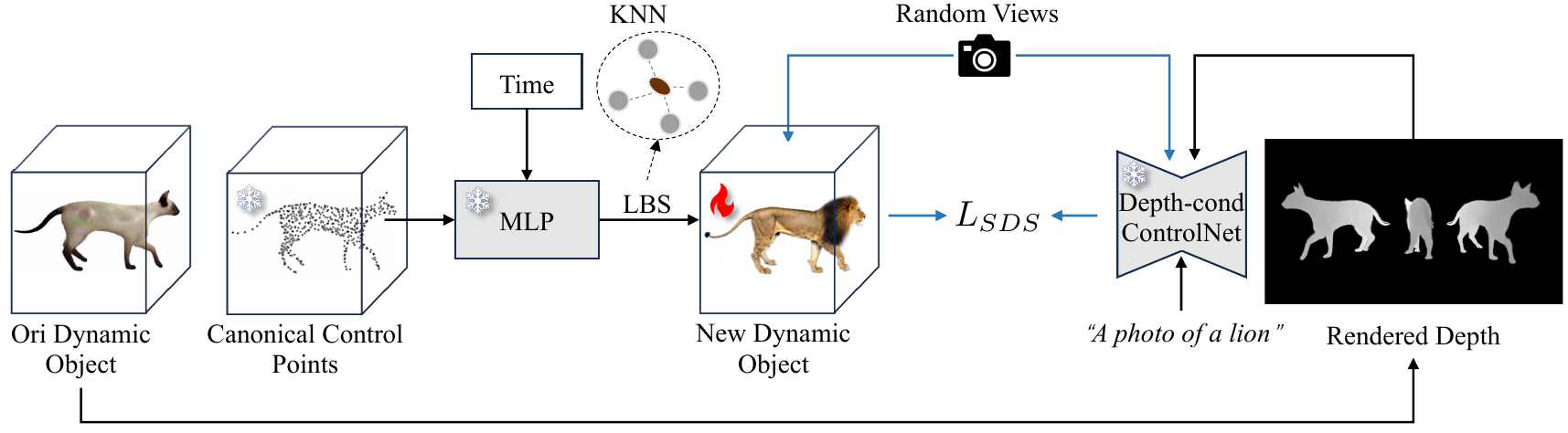}
\end{center}
% \vspace{-.1in}
\caption{Illustration of the pipeline for our motion transfer application.}
\label{fig:app}
\end{figure}

\subsection{Motion Transfer Application}
\label{motion_transfer}
Utilizing our video-to-4D framework, we can successfully extract the resultant dynamic 3D object as well as its motion represented by a set of moving control points. Leveraging this capability, we devise an application tailored for motion transfer predicated on the trajectories of these sparse control points. This application aims to synthesize dynamic objects of distinct entities that exhibit identical motion patterns, all instantiated through text descriptions.

As shown in Fig.~\ref{fig:app}, we employ the same initialization method as outlined in Sec.~\ref{fine}. During training, we fix the parameters of the learned control points as well as those of the deformation MLP, enabling us to preserve the motion of the target dynamic object and acquire its appearance from any viewpoint at any given time. To prevent the degeneration of motion and shape due to the displacement of dense Gaussians away from the control points during optimization, we elect to use a depth-condition ControlNet~\cite{controlnet} as supervision, and query the depth from the dynamic object resultant from the video-to-4D pipeline. Since no ground truth reference is involved, we only utilize SDS~\cite{dreamfusion} loss for optimization. We find that a small guidance scale can lead to plausible shape changes without degenerating the learned motion. Therefore, we first utilize a reduced guidance scale during the initial phase of the training regimen to capture the shape of the new entity that is congruent with the textual input, and its blurring appearance. Then, we save the intermediate dense Gaussians and employ the updated depth information as new conditions. In the remaining training period, we increase the guidance scale and decrease the noise scale of SDS to refine the texture, culminating in the final outcome. Please refer to Sec.~\ref{details} for more details.
\section{Experiments}
\label{exps}

\subsection{Experimental Settings}
\noindent\textbf{Implementation Details.}
\label{details}
We employ an MLP with a similar structure utilized in SC-GS~\cite{scgs} to predict the motion of control points. The MLP receives the positional embeddings~\cite{nerf} of time and control points' positions as input, with frequencies of 6 and 10. In the coarse stage, we initialize $M=512$ control Gaussians uniformly within a sphere with the same scaling parameter. We perform densification and pruning following 3DGS~\cite{3dgs} for the first 1,000 iterations with an interval of 100. Then we perform Farthest Point Sampling (FPS)~\cite{pointnet++} to sample $M$ control Gaussians and train for another 500 iterations without densification. In the fine stage, we optimize all the parameters of control points, MLP, and dense Gaussians together. During training, we sample 4 random camera poses and the fixed pose of time $t$ at a fixed radius of 2, with the azimuth in [-180, 180] degrees and elevation in [-$ver$, $ver$] degrees, where $ver=30$. We set the loss weight of $\lambda_{ref},\lambda_{mask},\lambda_{SDS},\lambda_{GA}$ to $5000.0,500.0,1.0,10000.0$ by default. As for our motion transfer application, during the initial 1,000 iterations, it is optimized with a guidance scale of 7.5 using the dynamic object's depth obtained from the video-to-4D pipeline as a condition. Subsequently, the depth from the intermediate result is utilized as the new condition, and we continue to optimize the remaining 1,000 iterations with a guidance scale of 30.0. We will release the source code later. 
% Please refer to \textit{Supp.} for more details.
Please refer to Sec.~1 of \textit{Supp.} for more details.

\noindent\textbf{Dataset.}
For fair comparisons, we utilize the dataset from Consistent4D~\cite{con4d} for evaluations. The dataset consists of 12 synthetic and 12 in-the-wild videos, all captured with a stationary camera oriented perpendicularly to the dynamic objects. Each video has 32 frames and spans around 2 seconds. 

\noindent\textbf{Evaluation Metrics.}
\label{metric}
We have summarized the criteria for evaluating the quality of video-to-4D generation into three main aspects: alignment with the reference video, spatio-temporal consistency, and motion fidelity. Following NeRF~\cite{nerf}, we utilize PSNR and SSIM~\cite{ssim} as measurements of the reference view alignment. We also add LPIPS~\cite{lpips} as a view quality metric, which is akin to Consistent4D~\cite{con4d}. As for multi-view (spatial) consistency, we adopt the commonly used CLIP~\cite{clip} score to measure the visual similarity of two different renderings. To evaluate the temporal consistency and motion fidelity of the generated dynamic object, we follow~\cite{ccpl} to utilize RAFT~\cite{raft} to compute the optical flows of the reference image sequence, and warp the reference-view projections to calculate the temporal error (Temp), which can effectively reflect the temporal consistency under the reference view and motion accuracy 
% (check \textit{Supp.} for details of the temporal error metric).
(check Sec.1 of \textit{Supp.} for details of the temporal error metric).
We also include human evaluation, which is more representative in generation tasks. To do so, we randomly choose 10 videos to train the compared methods and render the dynamic objects from the reference view and a random novel view. Then, we invite 20 participants to select their preferred reference and novel view videos based on the reference view alignment, spatio-temporal consistency, and motion fidelity. Overall, we get $10\times 20=200$ votes for reference view and novel view, respectively. Then, we calculate the percentage of votes as a measurement for user preference, as shown in Fig.~\ref{fig:user}.

\begin{figure}[!t]
\begin{center}
\includegraphics[width=0.98\textwidth]{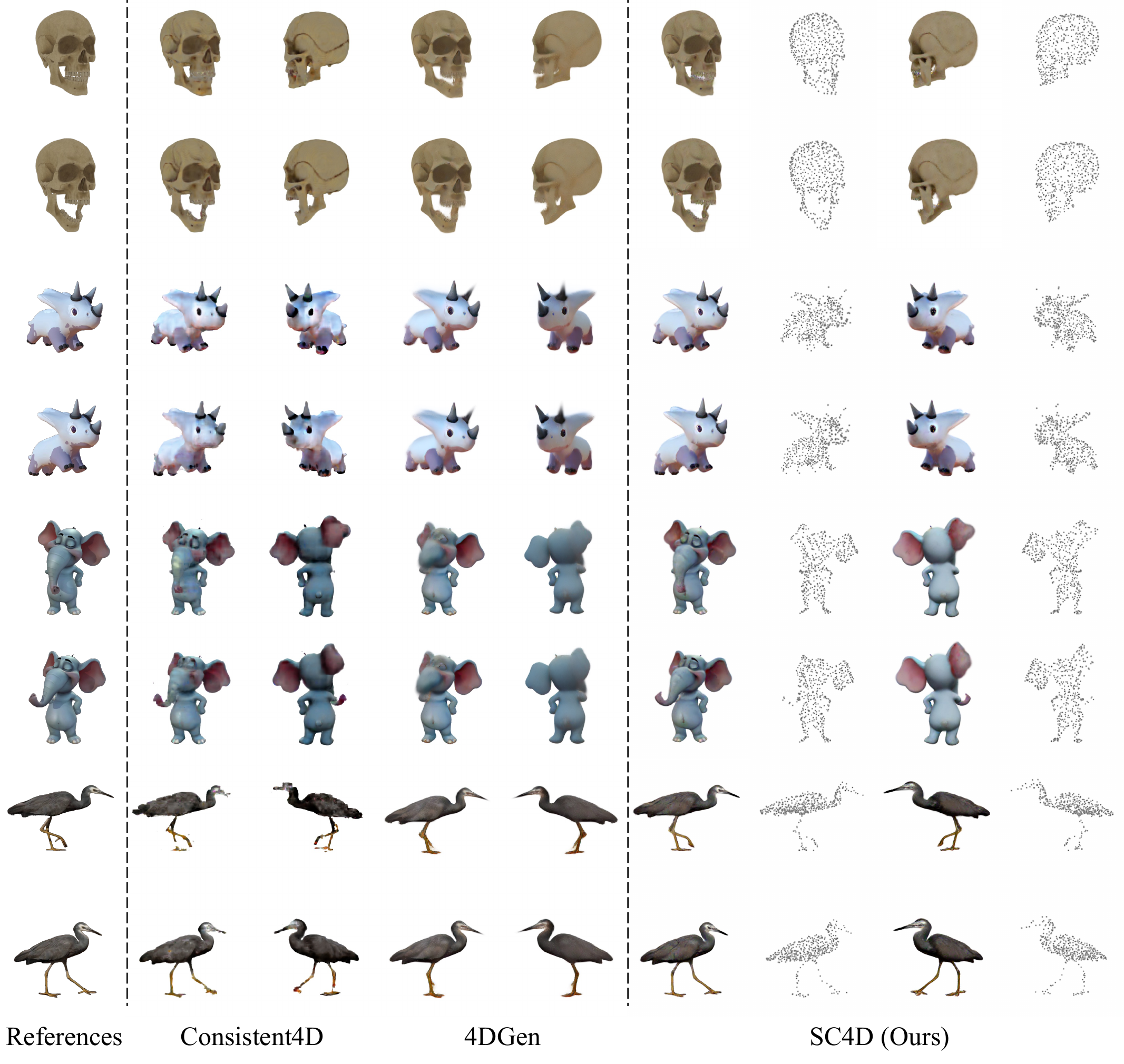}
\end{center}
%\vspace{-.2in}
\caption{Qualitative Comparisons. We compare our method with Consistent4D~\cite{con4d} and 4DGen~\cite{4dgen}. For each instance, we render two viewpoints at two timesteps. We also visualize the sparse control points to show their correspondence with dense Gaussians.}
\label{fig:comp}
\end{figure}

\subsection{Comparisons}
To evaluate the effectiveness of the proposed video-to-4D framework, we compare our method with the only two open-source methods: Consistent4D~\cite{con4d} and 4DGen~\cite{4dgen}. We train the dataset provided by Consistent4D with these methods and conduct comprehensive evaluations of the outcomes. All the results of Consistent4D and 4DGen are obtained using their official code and settings. Qualitative and quantitative comparisons are shown in Fig.~\ref{fig:comp} and Tab.~\ref{fig:comp}.

\noindent\textbf{Qualitative Comparison.}
As demonstrated in Fig.~\ref{fig:comp}, we show four instances (\textit{skull}, \textit{triceratops}, \textit{elephant}, \textit{egret}) and the results generated by the compared methods. Consistent4D~\cite{con4d} generates results with artifacts and color distortions in some cases (\textit{triceratops} and \textit{egret}), and there is also a noticeable temporal discontinuity between frames of the same viewpoint, which is attributable to the limited capacity of low-resolution NeRF~\cite{nerf} and the diversity of time-varying NeRF under single view conditions. As for 4DGen~\cite{4dgen}, the generated dynamic objects often exhibit minor motion and present discernible discrepancies from the reference video (\textit{skull} and \textit{triceratops}). Regions with large motion are subject to blurring or may even experience loss of detail (nose of \textit{elephant}, claw of \textit{egret}). In comparison, the results generated by our method surpass the compared methods in terms of alignment with the reference video, spatio-temporal consistency, and motion fidelity. This is also evidenced by the visualized sparse control points, from which it can be discerned that our method has learned a set of evenly distributed control points that accurately capture the dynamics and shape of the subject. 
% Please check \textit{Supp.} for more qualitative comparisons.   
Please check Sec.~5 of \textit{Supp.} for more qualitative comparisons.

\begin{table}[!t]
\centering
\caption{Quantitative comparison of different methods. Temp represents the temporal error introduced in Sec.~\ref{metric}.}
%\vspace{-.1in}
% \resizebox{0.9\linewidth}{!}{
\begin{tabular}{lcccccc}
\toprule[1pt] 
    Methods & PSNR~$\uparrow$ & SSIM~$\uparrow$ & LPIPS~$\downarrow$ & CLIP~$\uparrow$ & Temp~$\downarrow$ & Training time~$\downarrow$  \\
    \midrule 
    Consistent4D~\cite{con4d} & 23.97 & 0.91 & 0.09 & 0.89 & 0.0089 & 1.9h  \\
    4DGen~\cite{4dgen} & 21.80 & 0.90 & 0.10 & 0.87 & 0.0089 & 3.0h  \\
    \textbf{SC4D (Ours)} & \textbf{29.50} & \textbf{0.95} & \textbf{0.08} & \textbf{0.90} & \textbf{0.0081} & \textbf{1.0h}  \\
\bottomrule[1pt] 
\end{tabular} 
% }
%\vspace{-.1in}
\label{tab:quant}
\end{table}

\begin{figure}[t]
\begin{center}
\includegraphics[width=0.85\textwidth]{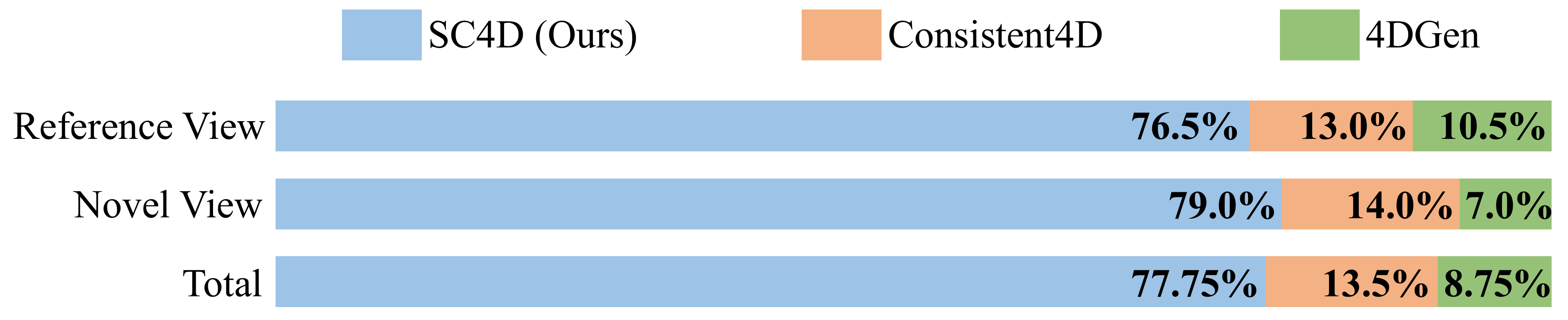}
\end{center}
%\vspace{-.2in}
\caption{User preference for video-to-4D generation methods.}
\label{fig:user}
%\vspace{-.21in}
\end{figure}

\noindent\textbf{Quantitative Comparison.}
To quantitatively evaluate the performance of the compared methods, we randomly choose 10 reference videos from the dataset provided by Consistent4D~\cite{con4d}, and optimize the dynamic objects using the compared methods, respectively. After optimization, we render one input view and four testing views (azimuth degrees: [72, 144, 216, 288], elevation degree: [0]) for every timestep of the reference video to calculate the metrics mentioned in Sec.~\ref{metric}. As reported in Tab.~\ref{tab:quant}, our method outperforms Consistent4D~\cite{con4d} and 4DGen~\cite{4dgen} in reference view alignment (PSNR, SSIM, LPIPS), multi-view consistency (CLIP), temporal consistency and motion fidelity (Temp), revealing the effectiveness of SC4D. As depicted in Fig.~\ref{fig:user}, user study also reveals that our proposed SC4D is more favored in human evaluations, in which the novel view preference metric also reveals the remarkable spatio-temporal consistency and motion fidelity of SC4D. Besides, our method exhibits a significant advantage in terms of training duration. All experiments are conducted on a single Tesla V100 GPU with 32 GB of graphics memory. We also utilize the test set and metrics in Consistent4D~\cite{con4d} to evaluate the compared methods. The results align with the conclusion drawn above. 
% We attach them in the \textit{Supp.}.
We attach them in Sec.~3 of \textit{Supp.}.

\begin{figure}[t]
\begin{center}
\includegraphics[width=0.95\textwidth]{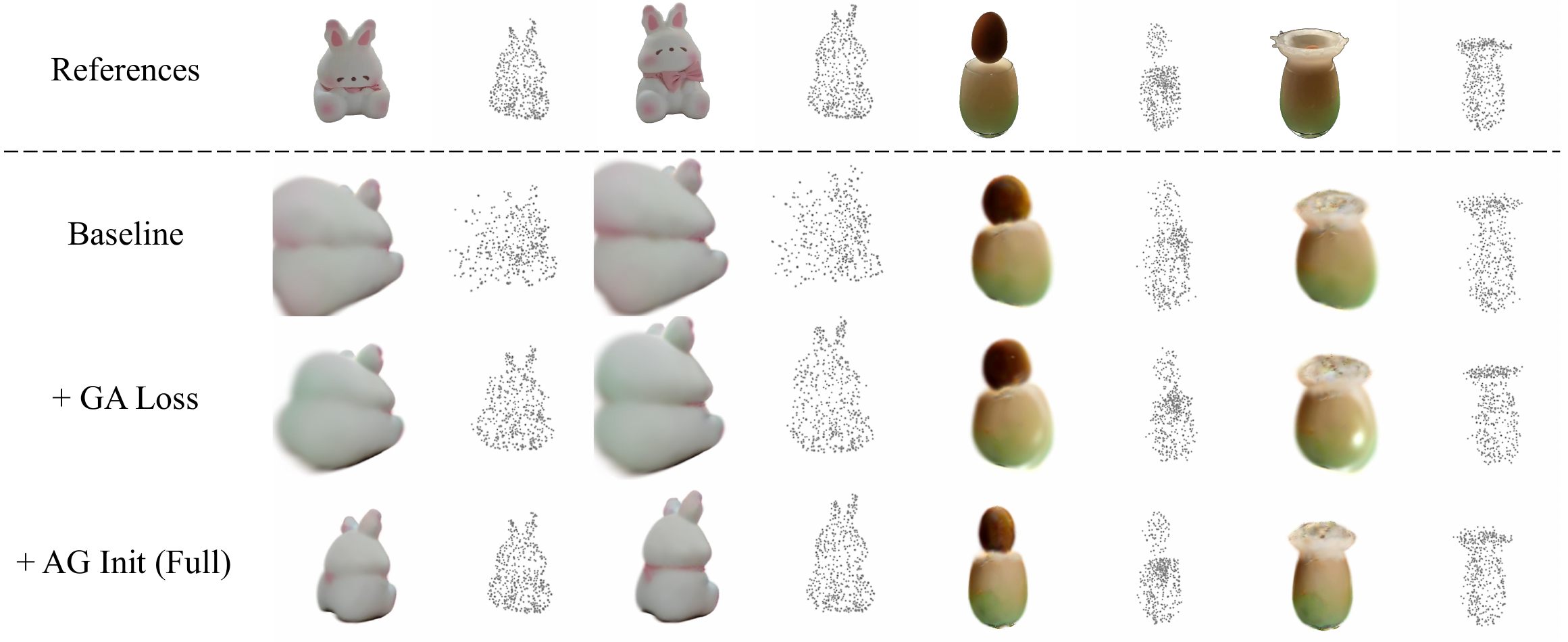}
\end{center}
%\vspace{-.15in}
\caption{Ablation studies of the proposed GA loss and AG initialization. In the first row, we show the source frames and control points from the coarse stage as references.}
\label{fig:abl}
\end{figure}

\subsection{Ablation Studies}
As mentioned in Sec.~\ref{intro}, we propose the Adaptive Gaussian (AG) initialization and Gaussian Alignment (GA) loss to alleviate the shape degeneration issue in the fine stage. As shown in Fig.~\ref{fig:abl}, without GA loss and AG initialization, the generated object is prone to suffer from over-thickness and texture blurring problems. Besides, the control points scatter, which indicates that the learned shape and motion in the coarse stage also degenerate significantly. When training with GA loss, the general shape of control points can be preserved, yet there can be observable instances of increased thickness. Additionally, when a viewpoint is excessively close to the camera, a corresponding attenuation in texture detail is also noticeable. Upon utilizing AG initialization, our method has experienced a pronounced enhancement in its final output. The dynamic objects generated exhibit tangible advancements in both motion fidelity and shape plausibility. Moreover, the textural details have also undergone considerable refinement.

\begin{table}[ht]
\centering
\caption{Quantitative evaluations of the proposed GA loss and AG initialization.}
%\vspace{-.1in}
%\resizebox{0.9\linewidth}{!}{
\begin{tabular}{lccccc}
\toprule[1pt] 
    Methods & PSNR~$\uparrow$ & SSIM~$\uparrow$ & LPIPS~$\downarrow$ & CLIP~$\uparrow$ & Temp~$\downarrow$   \\
    \midrule 
    Baseline & 29.81 & 0.95 & 0.10 & 0.82 & \textbf{0.0016}   \\
    + GA Loss & 30.19 & 0.96 & 0.09 & 0.83 & \textbf{0.0016}   \\
    \textbf{+ AG Init (Full)} & \textbf{31.35} & \textbf{0.96} & \textbf{0.08} & \textbf{0.89} & \textbf{0.0016}  \\
\bottomrule[1pt] 
\end{tabular} 
%}
\label{tab:abl}
\end{table}

Tab.~\ref{tab:abl} also illustrates the effectiveness of our proposed techniques. The most noticeable improvement in metrics is observed in the CLIP score, indicating a substantial enhancement in the quality of the generated novel views. This further reveals the effectiveness of the proposed GA loss and AG initialization in mitigating the challenges associated with shape degeneration. 
% Please refer to \textit{Supp.} for more ablation studies.  
Please refer to Sec.~4 of \textit{Supp.} for more ablation studies.

\begin{figure}[t]
\begin{center}
\includegraphics[width=0.95\textwidth]{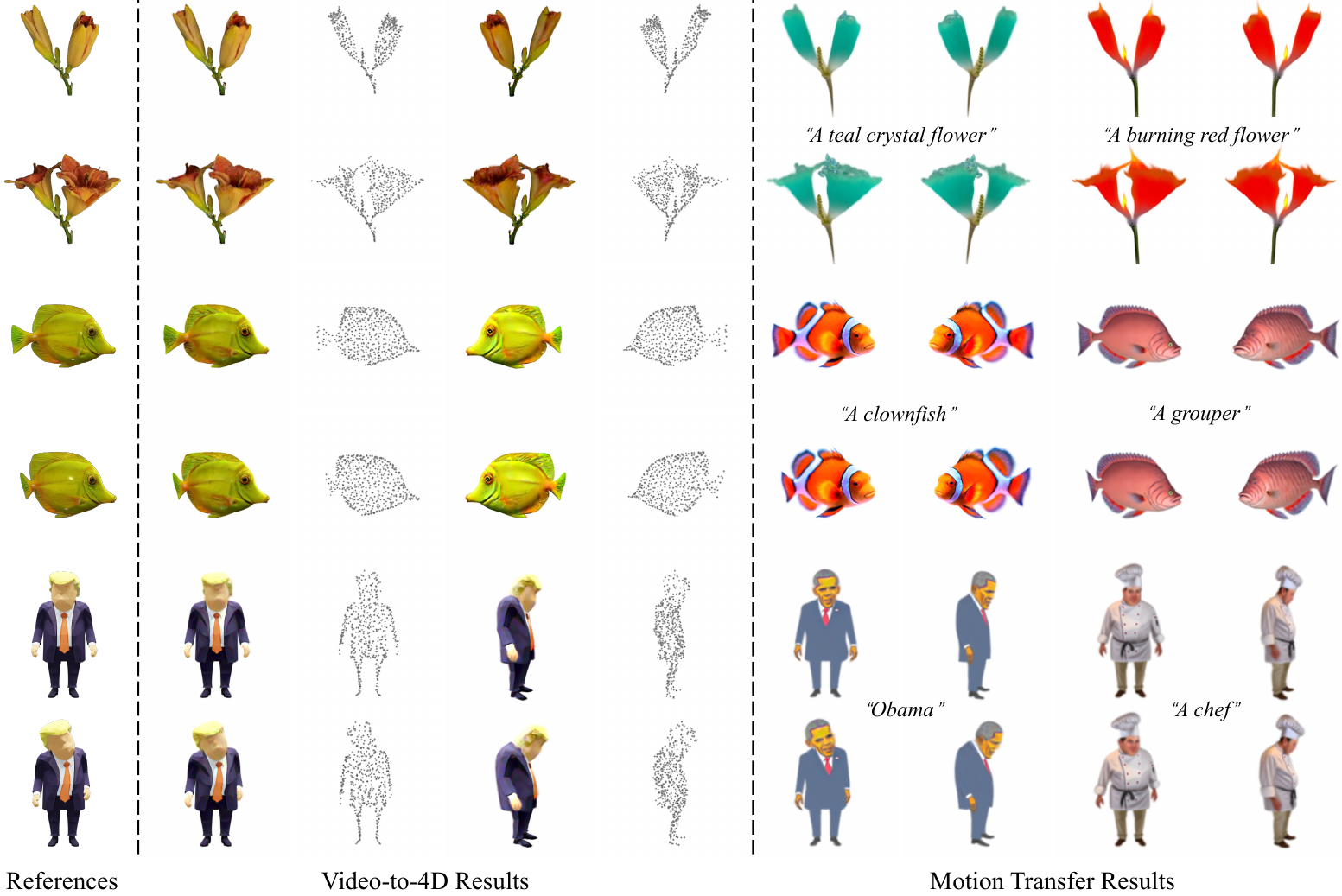}
\end{center}
%\vspace{-.15in}
\caption{Motion transfer application of SC4D. The prompt is attached between lines.}
\label{fig:app_qua}
\end{figure}

\subsection{Application Results}
As mentioned in Sec.~\ref{motion_transfer}, we design an innovative application that is capable of flexibly generating diverse 4D entities with the same motion, based on control points learned during our video-to-4D pipeline, in conjunction with a depth-condition ControlNet~\cite{controlnet} that operates upon textual descriptions.

In Fig.~\ref{fig:app_qua}, we sequentially present: the reference frames, video-to-4D results generated using SC4D and corresponding control points visualizations, and the new entities with identical motion synthesized based on text descriptions. From the figure, it is observable that the newly generated 4D objects possess vivid textures and align well with the motion patterns of the reference video. Moreover, the synthesized dynamic objects exhibit appreciable variations in shape, yet the motion remains coherent, which further illustrates the robustness and flexibility of our method. 
% Please refer to \textit{Supp.} for more application examples.
Please refer to Sec.~6 of \textit{Supp.} for more application examples.
\section{Limitations}
\label{limits}

The main limitations of our method are twofold: First, our approach relies on models such as Zero123~\cite{zero123} to provide novel view information, and the capability of these viewpoint synthesis models is still limited, often underperforming on many complex objects. Second, similar to existing video-to-4D methods~\cite{con4d, 4dgen, efficient4d}, we have not taken into account the 4D generation under moving camera scenarios, which will be one of the directions for our future research.
\section{Conclusion}
\label{conclusion}

In this work, we propose a sophisticated video-to-4D pipeline, named SC4D, which decouples the motion and appearance of dynamic objects as sparse control points and dense 3D Gaussians. SC4D excels over contemporary approaches in generating dynamic objects with better reference view alignment, spatio-temporal consistency, and motion fidelity. Moreover, we craft an innovative application utilizing the control points learned by SC4D, which allows seamless motion transfer onto new entities, as directed by textual descriptions.

\section*{Acknowledgements}
% This work was supported by the National Science Fund for Distinguished Young Scholars of China (Grant No.62225603).
This work was supported by Damo Academy through Damo Academy Research Intern Program.

\bibliographystyle{splncs04}
\bibliography{main}
\end{document}